\documentclass[letterpaper]{article} % DO NOT CHANGE THIS
\usepackage{aaai2026}  % DO NOT CHANGE THIS
\usepackage{times}  % DO NOT CHANGE THIS
\usepackage{helvet}  % DO NOT CHANGE THIS
\usepackage{courier}  % DO NOT CHANGE THIS

\usepackage[hyphens]{url}  % DO NOT CHANGE THIS
\usepackage{graphicx} % DO NOT CHANGE THIS
\usepackage{natbib}  % DO NOT CHANGE THIS AND DO NOT ADD ANY OPTIONS TO IT
\usepackage{caption} % DO NOT CHANGE THIS AND DO NOT ADD ANY OPTIONS TO IT
\usepackage{algorithm}
\usepackage{algorithmic}
\usepackage{amssymb}
\usepackage{booktabs}
\usepackage[utf8]{inputenc}
\usepackage{amsmath}
\usepackage{enumitem}
\usepackage{newfloat}
\usepackage{makecell}
\usepackage{placeins}
\newcommand{\equalcontrib}{}
\usepackage{listings}
\DeclareCaptionStyle{ruled}{labelfont=normalfont,labelsep=colon,strut=off} % DO NOT CHANGE THIS

\newfloat{listing}{tb}{lst}{}
\floatname{listing}{Listing}
\title{ViInfographicVQA: A Benchmark for Single and \MultiImage Visual Question Answering on Vietnamese Infographics}

\author{
    Tue-Thu Van-Dinh\textsuperscript{\rm 1}\equalcontrib, 
    Hoang-Duy Tran\textsuperscript{\rm 1}\equalcontrib, 
    Truong-Binh Duong\textsuperscript{\rm 1}, 
    Mai-Hanh Pham\textsuperscript{\rm 1}, \\
    Binh-Nam Le-Nguyen\textsuperscript{\rm 1}, 
    Quoc-Thai Nguyen\textsuperscript{\rm 1,2}\thanks{Corresponding author: \texttt{quocthai@neu.edu.vn}.\\ We thank AI VIETNAM for facilitating computational resources and financial support for this paper.}
}

\affiliations{
    \textsuperscript{\rm 1}AI VIETNAM Lab, Vietnam\\
    \textsuperscript{\rm 2}National Economics University, Vietnam
}

\usepackage{bibentry}

\newcommand{\NImg}{\the\numexpr \NStr+\NMtr+\NSte-\NOverlapTr-\NOverlapTe+\NMte\relax}
\newcommand{\NQA}{\the\numexpr \QSte+\QMte+\QStr+\QMtr\relax}
\newcommand{\NStr}{1788}\newcommand{\NSte}{192}
\newcommand{\QStr}{12521}\newcommand{\QSte}{1374}
\newcommand{\NMtr}{4823}\newcommand{\NMte}{509}
\newcommand{\QMtr}{5878}\newcommand{\QMte}{636}
\newcommand{\NMGrpTr}{2090}\newcommand{\NMGrpTe}{218}
\newcommand{\NOverlapTr}{515}\newcommand{\NOverlapTe}{50}

\newcommand{\ARmin}{0.33}   % min aspect ratio (w/h)
\newcommand{\ARmax}{3.00}   % max aspect ratio (w/h)
\newcommand{\MinSide}{512}  % min shorter side in pixels

\usepackage{xspace}
\usepackage{enumitem}

\newcommand{\SingleImage}{Single-image\xspace}
\newcommand{\MultiImage}{Multi-image\xspace}
\newcommand{\ImageSpan}{Image-span\xspace}
\newcommand{\QuestionSpan}{Question-span\xspace}
\newcommand{\MultiSpan}{Multi-span\xspace}
\newcommand{\NonExtractive}{Non-extractive\xspace}
\newcommand{\CrossImageSynthesis}{Cross-image synthesis\xspace}
\newcommand{\NonSpan}{Non-span\xspace}
\newcommand{\MultiImageSpan}{Multi-image span\xspace}

\newcommand{\term}[1]{\emph{#1}}

\newcommand{\best}[1]{\textbf{#1}}
\newcommand{\second}[1]{\underline{#1}}

\begin{document}

\maketitle
\begin{abstract}
Infographic Visual Question Answering (InfographicVQA) evaluates a model’s ability to read and reason over data-rich, layout-heavy visuals that combine text, charts, icons, and design elements. Compared with scene-text or natural-image VQA, infographics require stronger integration of OCR, layout understanding, and numerical and semantic reasoning. We introduce ViInfographicVQA, the first benchmark for Vietnamese InfographicVQA, comprising over \NImg{} real-world infographics and \NQA{} human-verified question-answer pairs across economics, healthcare, education, and more. The benchmark includes two evaluation settings. The \SingleImage task follows the traditional setup in which each question is answered using a single infographic. The \MultiImage task requires synthesizing evidence across multiple semantically related infographics and is, to our knowledge, the first Vietnamese evaluation of cross-image reasoning in VQA. We evaluate a range of recent vision-language models on this benchmark, revealing substantial performance disparities, with the most significant errors occurring on \MultiImage questions that involve cross-image integration and non-span reasoning. ViInfographicVQA contributes benchmark results for Vietnamese InfographicVQA and sheds light on the limitations of current multimodal models in low-resource contexts, encouraging future exploration of layout-aware and cross-image reasoning methods.
\end{abstract}

% \begin{abstract}
% Infographic Visual Question Answering (InfographicVQA) evaluates a model’s ability to read and reason over data-rich, layout-heavy visuals that combine text, charts, icons, and design elements. Compared with scene-text or natural-image VQA, infographics require stronger integration of OCR, layout understanding, and numerical and semantic reasoning. We introduce ViInfographicVQA, the first benchmark for Vietnamese InfographicVQA, comprising over \NImg{} real-world infographics and \NQA{} human-verified question–answer pairs across diverse domains such as economics, healthcare, and education. The benchmark includes two evaluation settings: the \SingleImage task, where each question is answered using a single infographic, and the \MultiImage task, which requires synthesizing evidence across multiple semantically related infographics-the first such evaluation of cross-image reasoning in Vietnamese VQA. Benchmarking recent vision–language models reveals substantial performance gaps, especially on \MultiImage questions that require synthesizing information across images and performing non-extractive reasoning. ViInfographicVQA provides the first benchmark results for Vietnamese infographic understanding and highlights key limitations of current multimodal models in low-resource contexts, motivating future exploration of layout-aware and cross-image reasoning approaches.
% \end{abstract}

\begin{links}
    \link{Code}{https://github.com/duongtruongbinh/ViInfographicVQA}
    \link{Datasets}{https://huggingface.co/datasets/duytranus/ViInfographicVQA}
    % \link{Extended version}{https://aaai.org/example/extended-version}
\end{links}

\section{Introduction}
Despite substantial progress in multimodal reasoning, most Visual Question Answering (VQA) benchmarks remain focused on English and emphasize natural images or simple   

\begin{figure}[H]
    \centering
    \includegraphics[width=\linewidth]{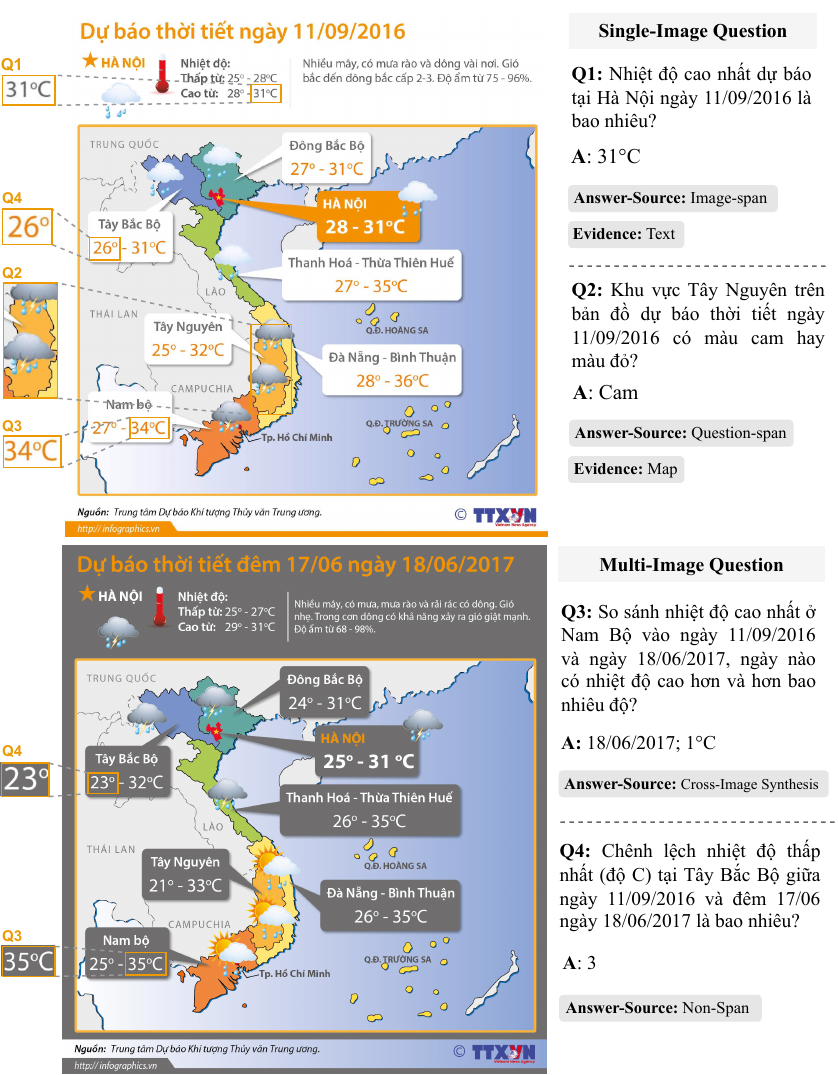}
    \caption{Example images from \textbf{ViInfographicVQA} with questions and answers. Each question is annotated with its answer source, supporting evidence type, and required operation.}
    \label{fig:sample}
\end{figure}

\begin{figure*}[h]  
    \centering
    \includegraphics[width=1\linewidth]{images/single_multi_sample2.pdf}
    \caption{Representative examples illustrating the diversity of \textbf{ViInfographicVQA}. The dataset covers a wide range of infographic types (e.g., bar charts, energy reports, event schedules) and question forms spanning extractive, non-extractive, and \MultiImage reasoning. This diversity enables comprehensive evaluation of multimodal models' abilities in text understanding, numerical reasoning, and cross-image aggregation.}
    \label{fig:enter-label}
\end{figure*}

\noindent scene-text settings \cite{antol2015vqa}. Such datasets rarely capture the complexity of structured, data-rich visuals like \term{infographics}, which blend text, charts, icons, and layout for concise information delivery. Understanding these materials requires unified reasoning over visual, textual, and spatialmodalities - a challenge extending beyond typical image-based VQA.

Infographics have recently attracted interest in document-oriented VQA, where models must comprehend layout, embedded text, and numerical data simultaneously \citep{mathew2021docvqa,mathew2022infographicvqa,masrychartqa}. However, existing benchmarks remain largely English-centric and provide limited evaluation of layout-aware or numerically grounded reasoning \citep{singh2019towards,biten2019scene,tran2021vivqa,doan2024vintern1b}. While InfographicVQA \citep{mathew2022infographicvqa} established the first English dataset for infographic understanding, comparable resources in low-resource languages such as Vietnamese are still absent-particularly for tasks requiring reasoning across multiple infographics.

In the Vietnamese context, prior datasets such as ViVQA \cite{tran2021vivqa}, EVJVQA \cite{Luu_Thuy_Nguyen_2023}, OpenViVQA \cite{nguyen2023openvivqa}, and ViTextVQA \cite{van2024vitextvqa} provide valuable contributions but mainly target natural images, OCR scene text, or synthetic renderings. None address the more complex domain of infographic understanding or \term{cross-image reasoning}, which demands synthesis of information across multiple semantically related visuals.

To bridge this gap, we introduce \textbf{ViInfographicVQA}-a Vietnamese benchmark designed to evaluate both \SingleImage comprehension and \MultiImage reasoning. The dataset features linguistically authentic questions and answers across diverse topics such as economics, healthcare, and education, offering two complementary evaluation settings (Figure~\ref{fig:sample}).

\noindent Our contributions are summarized as follows:
\begin{enumerate}[label=(\roman*), leftmargin=*, itemsep=2pt, topsep=2pt]
    \item We introduce \textbf{ViInfographicVQA}, a Vietnamese infographic VQA dataset with official splits, containing over \NImg{} infographics and nearly \NQA{} QA pairs.
    \item We design both \SingleImage and \MultiImage tasks; the latter enforces cross-image evaluation previously unaddressed in Vietnamese or multilingual VQA.
    \item We establish baselines with recent vision-language models (VLMs) and conduct systematic fine-tuning on subsets and the combined data, analyzing training strategies without rationale leakage.
\end{enumerate}

\begin{table*}[!h]
    \centering
    \resizebox{\textwidth}{!}{%
    \begin{tabular}{@{}llcclrrl@{}}
        \toprule
        \textbf{Dataset} & \textbf{Image Source} & \textbf{Synthetic Images} & \textbf{Template Questions} & \textbf{Text Type} & \textbf{\# Images} & \textbf{\# Questions} & \textbf{Answer Type} \\ 
        \midrule
        VQA v1/v2 & Natural images & $\times$ & $\checkmark$ & MR & 200K+ & 1M+ & MCQ \\
        ST-VQA & Natural images & $\times$ & $\times$ & ST & 28K & 45K & Ex \\
        TextVQA & Natural images & $\times$ & $\times$ & ST & 28K & 45K & Ex, SAb \\
        DocVQA & Industry documents & $\times$ & $\times$ & Pr, Tw, Hw, BD & 12K & 50K & Ex \\
        VisualMRC & Webpage screenshots & $\times$ & $\times$ & BD & 10K & 30K & Ab \\
        InfographicVQA & Infographics & $\times$ & $\times$ & BD & 5.4K & 30K & Ex, Nm \\
        FigureQA & Charts (synthetic) & $\checkmark$ & $\checkmark$ & BD & 120K & 1.5M & Y/N \\
        DVQA & Bar charts & $\checkmark$ & $\checkmark$ & BD & 300K & 3.4M & Ex, Nm, Y/N \\
        PlotQA & Scientific plots & $\checkmark$ & $\checkmark$ & BD & 224K & 28.9M & Ex \\
        ChartQA & Real charts & $\times$ & $\times$ & BD & 20K & 400K & Ex \\
        CharXiv & Scientific charts (arXiv) & $\times$ & $\times$ & BD & 10K+ & 200K+ & Ex, Ab \\
        InfoChartQA & Infographic charts & $\times$ & $\times$ & BD & 15K & 300K & Ex, Ab \\
        ChartQAPro & Diverse charts & $\times$ & $\times$ & BD & 50K & 1M+ & Ex, Ab, Nm \\
        ChartGalaxy & Mixed infographic charts & $\checkmark$ & $\times$ & BD & 1M+ & 20M+ & Ex, Ab \\
        \midrule
        ViVQA & MS COCO images & $\times$ & $\times$ & ST & 10.3K & 15K & Ex \\
        ViTextVQA & Natural images with text & $\times$ & $\times$ & ST & 16.8K & 50.3K & Ex \\
        OpenViVQA & Web images of Vietnamese life & $\times$ & $\times$ & ST & 11.2K & 37.9K & Ab \\
        ViCLEVR & Rendered CLEVR images & $\checkmark$ & $\times$ & ST & 26K & 30K & Ex / SAb \\
        ViOCRVQA & Book covers & $\times$ & $\checkmark$ & BD & 28.3K & 121.6K & Ex \\
        EVJVQA & Web images of Vietnamese life & $\times$ & $\times$ & ST & 4.9K & 33.8K & Ex, Ab \\
        Viet-OCR-VQA & Scene text (Vietnamese) & $\times$ & $\times$ & ST & 137K & 822K+ & Ex \\
        Viet-Doc-VQA & Textbooks, documents & $\times$ & $\times$ & Pr, BD & 116K+ & 500K+ & Ex \\
        Viet-Handwriting-VQA & Handwritten text & $\times$ & $\times$ & Hw & 20K+ & 100K+ & Ex \\
        Viet-Receipt-VQA & Receipts & $\times$ & $\times$ & Pr, BD & 30K+ & 200K+ & Ex \\
        Viet-Menu-VQA & Menus & $\times$ & $\times$ & Pr, BD & 10K+ & 50K+ & Ex \\
        \bottomrule
    \end{tabular}%
    }
    \caption{
        A comparative overview of VQA datasets. This table lists key attributes, including image source, use of synthetic images or template-based questions, the number of images and questions, along with the supported text types and answer formats.
        \textbf{Text Type}: MR (Multimodal Reading), ST (Scene Text), BD (Born-Digital), Pr (Printed), Tw (Typewritten), Hw (Handwritten). 
        \textbf{Answer Type}: MCQ (Multiple Choice Question), Ex (Extractive), SAb (Short Abstractive), Ab (Abstractive), Nm (Numeric), Y/N (Yes/No).
    }
    \label{tab:vqa_datasets}
\end{table*}

\section{Related Work}
\subsection{Preceding Visual Question Answering Datasets}
Early benchmarks such as VQA v1 \cite{antol2015vqa} and VQA v2 \cite{goyal2017making} established the task but primarily evaluate models on natural images. Subsequent studies show that general-purpose VQA systems struggle when reading text or interpreting the layout is required. To probe these abilities, text- and document-centric datasets were introduced. ST-VQA \cite{biten2019scene} and TextVQA \cite{singh2019towards} target scene-text reading and reasoning. DocVQA \cite{mathew2021docvqa} extends the evaluation to scanned business and administrative documents, and VisualMRC \cite{tanaka2021visualmrc} focuses on webpage screenshots. More recently, PDF-VQA \cite{zhang2023pdfvqa} considers complex multi-page PDFs with diverse layouts.

A second line of research focuses on data visualizations. Synthetic testbeds such as FigureQA \cite{kahou2017figureqa} and DVQA \cite{kafle2018dvqa} allow controlled analysis but lack realism. PlotQA \cite{methani2020plotqa} scales to millions of QA pairs over scientific plots. ChartQA \cite{masrychartqa} introduces real-world charts with human-authored questions, while CharXiv \cite{wang2024charxiv} curates scientific charts from arXiv. More recent efforts-InfoChartQA \cite{lin2025infochartqa}, ChartQAPro \cite{masry2025chartqapro}, and ChartGalaxy \cite{li2025chartgalaxy}-explicitly include infographic-style charts and expand both coverage and diversity.

\subsection{Infographic Question Answering Benchmarks}
Infographics combine textual, numerical, and graphical information within structured layouts, posing distinct challenges that differ from those in natural images, scene text, or document-based VQA. To address this complexity, several benchmarks have been developed to specifically evaluate reasoning over infographic-style content. InfographicVQA \cite{mathew2022infographicvqa} introduced the first dataset in this domain, featuring questions that require textual understanding and reasoning over layout and visual design. Building on this direction, InfoChartQA \cite{lin2025infochartqa} investigates how visual elements such as icons and pictograms influence question answering, ChartQAPro \cite{masry2025chartqapro} expands chart types and question formats while explicitly covering infographic-style data, and ChartGalaxy \cite{li2025chartgalaxy} scales the scope to millions of real and synthetic infographic charts. Despite these advances, existing benchmarks remain predominantly English-centric, leaving non-English infographic reasoning largely unexplored.

\begin{figure*}[h] 
    \centering
    \includegraphics[width=1\linewidth]{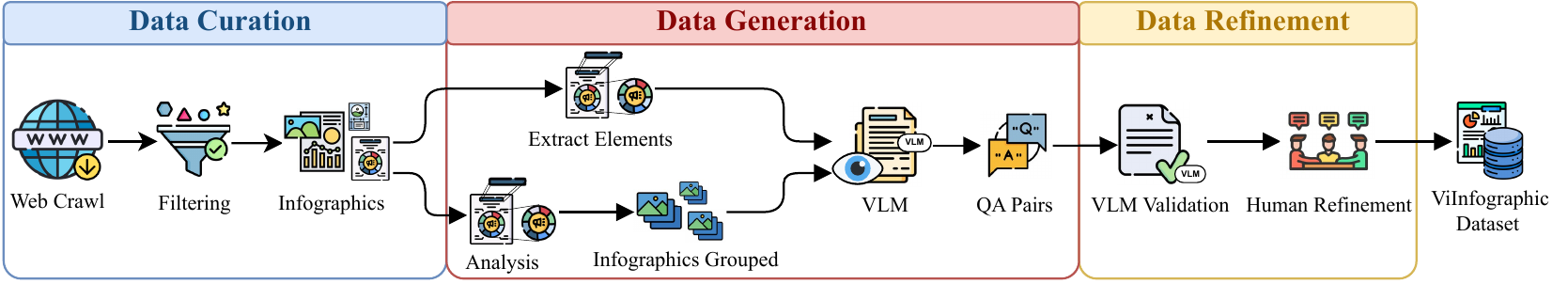}
\caption{
Construction pipeline for \textbf{ViInfographicVQA}.
\textbf{Data curation} crawls the web, filters infographics, and selects high-quality ones for QA construction.
\textbf{Data generation} uses a two-branch VLM pipeline: element extraction from single images for factoid QAs, and grouping related infographics to elicit cross-image questions.
\textbf{Data refinement} applies automated VLM validation followed by human review, yielding the final dataset.
}
    \label{fig:pipeline}
\end{figure*}

\subsection{Visual Question Answering Datasets in Vietnamese}
Vietnamese VQA has advanced along several directions but largely outside the infographic domain. ViVQA \cite{tran2021vivqa} adapts a subset of VQA v2, with limited linguistic naturalness. EVJVQA \cite{Luu_Thuy_Nguyen_2023} provides multilingual QA pairs but lacks scene text. OpenViVQA \cite{nguyen2023openvivqa} supports open-ended answers, and ViCLEVR \cite{tran2024viclevr} targets compositional reasoning. ViTextVQA \cite{van2024vitextvqa} is the first Vietnamese scene-text dataset, while ViOCRVQA \cite{pham2025viocrvqa} scales OCR-based QA. The Vintern-1B collection \cite{doan2024vintern1b} adds Vietnamese resources for OCR, document understanding, handwriting, and extraction (e.g., Viet-OCR-VQA, Viet-Doc-VQA, Viet-Handwriting-VQA, Viet-Receipt-VQA; many are hosted by 5CD-AI on HuggingFace\footnote{\url{https://huggingface.co/5CD-AI}}). These datasets lay crucial groundwork yet remain limited for infographic understanding. Many rely on translated English data or synthetic renderings, whereas our work leverages authentic Vietnamese infographics with diverse layouts and topics. Table~\ref{tab:vqa_datasets} consolidates English and Vietnamese resources and highlights the lack of a Vietnamese benchmark for infographic VQA-especially one that evaluates cross-image (\MultiImage) reasoning.

\section{ViInfographicVQA}
The creation of \textbf{ViInfographicVQA} begins by sourcing and filtering real-world Vietnamese infographics, which are then organized into semantically related sets to support \MultiImage reasoning. Figure~\ref{fig:enter-label} presents representative examples from the dataset, illustrating the diversity of question types and reasoning operations covered across both \SingleImage and \MultiImage tasks. Figure~\ref{fig:pipeline} then outlines the end-to-end construction process, combining VLM-assisted element extraction, clustering, and structured QA generation with subsequent human verification and refinement to produce high-quality, linguistically natural question–answer pairs.

\subsection{Data Curation}
We curate real-world Vietnamese infographics and organize them into semantically related sets to support both \SingleImage and \MultiImage reasoning. The process has 3 steps: sourcing and geometry filtering, VLM-assisted representation and grouping, and split assignment.

\textit{Step 1: Sourcing and geometry filtering.}
We collect infographics from the portal \textit{infographics.vn}\footnote{\url{https://infographics.vn/}, a data visualization and infographics portal maintained by the Vietnam News Agency (VNA), the primary and official news provider of the Vietnamese government.} across topics such as Economics, Healthcare, Culture-Society, Disaster-Accident, and Sports-Arts. To ensure OCR and layout parsing readability while avoiding extreme rescaling artifacts, we retain images whose aspect ratios fall within $[\ARmin{},\ARmax{}]$ and whose shorter side is at least $\MinSide{}$ pixels.

\textit{Step 2: VLM-assisted representation and grouping.}
To capture relationships among infographics, we compute multimodal representations combining OCR text features with visual and layout cues. Concretely, we extract (i) normalized OCR text and its locations, (ii) global and region-level visual descriptors, and (iii) lightweight layout signals (e.g., panel/legend cues) using a VLM-assisted pass that proposes informative regions and aggregates them into image-level embeddings. We cluster images within topical categories using cosine similarity with a fixed number of clusters per category (3 per topic), forming semantically coherent sets that enable cross-image reasoning.

\textit{Step 3: Split assignment with set isolation.}
Entire \MultiImage sets are assigned to a single split (train/val/test) to prevent cross-split leakage. Splits are further stratified by (topic $\times$ answer-source) priors to preserve distributional balance for downstream evaluation.

\subsection{Data Annotation}
\subsubsection{\SingleImage Infographic}
Infographics combine text blocks, tables, charts, maps, and icons; these elements are natural anchors for structured QA. Rather than treating an infographic as a single visual unit, we first identify semantically meaningful regions and then build QA on top of them.

\textit{Stage 1: Element extraction.}
A VLM-assisted scan proposes informative regions (text boxes, chart panels, legends, icons, map areas) and refines their coordinates using visual and spatial cues to preserve contextually relevant content.

\textit{Stage 2: QA construction.}
The extracted elements serve as the semantic backbone for question generation, especially for reasoning over layout, text, graphics, and data visualizations~\cite{mathew2022infographicvqa}. We organize QA pairs along two independent dimensions:
(i) \emph{element types} (e.g., table, graph/diagram, map, text, layout/visual), and
(ii) \emph{answer-source categories}:
\ImageSpan\ (a single, contiguous verbatim span in the image),
\QuestionSpan\ (multiple-choice with an explicit option list; the answer is exactly one option, verbatim),
\MultiSpan\ (two or more non-contiguous verbatim spans; the answer is a single string formed by joining the selected spans with exactly the comma delimiter between adjacent spans; the question is a single sentence specifying the task requirements and the properties of the spans), and
\NonExtractive\ (numeric reasoning such as counting, arithmetic, or computing a rank/percentage; the final output is a single numeric value).
To instantiate diverse QAs, we use rule-based prompt templates that pair element types with answer-source categories (e.g., \textit{Table} $\times$ \MultiSpan: “Which rows have values exceeding 100?”; \textit{Graph} $\times$ \NonExtractive: “What is the growth rate between the two years?”).

\noindent Unlike InfographicVQA~\cite{mathew2022infographicvqa}, we do not maintain separate labels for operations (counting, arithmetic, sorting). All numeric reasoning is grouped into \NonExtractive. Comparative behaviors that may occur in \ImageSpan, \QuestionSpan, or \MultiSpan\ (e.g., picking a maximum, minimum, or top-$k$) are considered inherent to the question semantics, not additional operation tags.

\textit{Stage 3: Verification.}
We employ VLM-assisted expansion (Gemini 2.0 Flash \cite{gemini20flash}) to generate multiple candidate QAs per rule, retaining short validation explanations but releasing answers only to avoid rationale leakage. Human reviewers then filter obvious knowledge questions (e.g. questions answerable without reference to the image), refine wording, and ensure linguistic naturalness and semantic faithfulness. 

\noindent\emph{Annotation Tool.} Gemini~2.0~Flash is used solely for semi-automatic dataset annotation during QA pair generation; it produces candidate QAs and brief explanations for internal validation only. Answers (without rationales) are released, and Gemini is not included among the evaluated baselines.

\subsubsection{\MultiImage Infographics}
Many real scenarios require synthesis across related visuals. Our \MultiImage annotation follows the same three-stage recipe, applied to semantically grouped infographics.

\textit{Stage 1: Set definition.}
We construct sets of related images and define the relation linking them (layout similarity, subject consistency, content variation, shared theme, or differing emphasis). Each set is assigned one of four relation types:
(i) \textit{Identical Replication} (same template, subject, and content);
(ii) \textit{Content Augmentation/Variation within the Same Specific Subject} (same template/subject, varying content);
(iii) \textit{Content Replicated, Presentation Varies} (same subject/content, different templates);
(iv) \textit{Thematic Connection with Diverse Specific Subjects/Focus} (different specific subjects under a common theme).

\textit{Stage 2: QA construction.}
Prompt templates instruct VLM-assisted generation to create questions that \emph{require} cross-image evidence. We categorize outputs as:
\MultiImageSpan\ (aggregate spans from $\ge$ 2 images under a logical condition),
\CrossImageSynthesis\ (compare/interpret across images to reach a new conclusion),
and \NonSpan\ (numerical computation such as sum, difference, average, or counting over values drawn from different images).

\textit{Stage 3: Verification.}
An automated checker removes duplicates, tests answer-explanation consistency, and verifies that each question genuinely depends on at least two images. Items flagged by the checker are reviewed and corrected by human annotators. As in the \SingleImage case, we adopt VLM-assisted expansion with explanations used only for internal validation and do not release rationales.

For transparency and reproducibility, the exact prompt templates used to construct all QA pairs are listed in Appendix~A.

\subsection{Data Refinement}
We apply a multi-step refinement process combining automated validation with human review \cite{baechler2024screenai, ding2025building} to ensure data quality and reliability. First, \emph{deduplication} is performed to remove redundant or near-duplicate QA pairs, preserving diversity and preventing overfitting to repeated patterns. Next, \emph{automated validation} with VLMs verifies that generated answers are consistent with infographic content, flagging potential errors or inconsistencies. Finally, \emph{human review} on a web platform edits question phrasing for clarity, resolves ambiguities, and ensures that answers are contextually aligned with infographic semantics. By integrating automated validation with human review, this human-in-the-loop approach ensures an accurate, linguistically natural, and semantically coherent dataset.

\begin{figure}[h]
    \centering
    \includegraphics[width=1\linewidth]{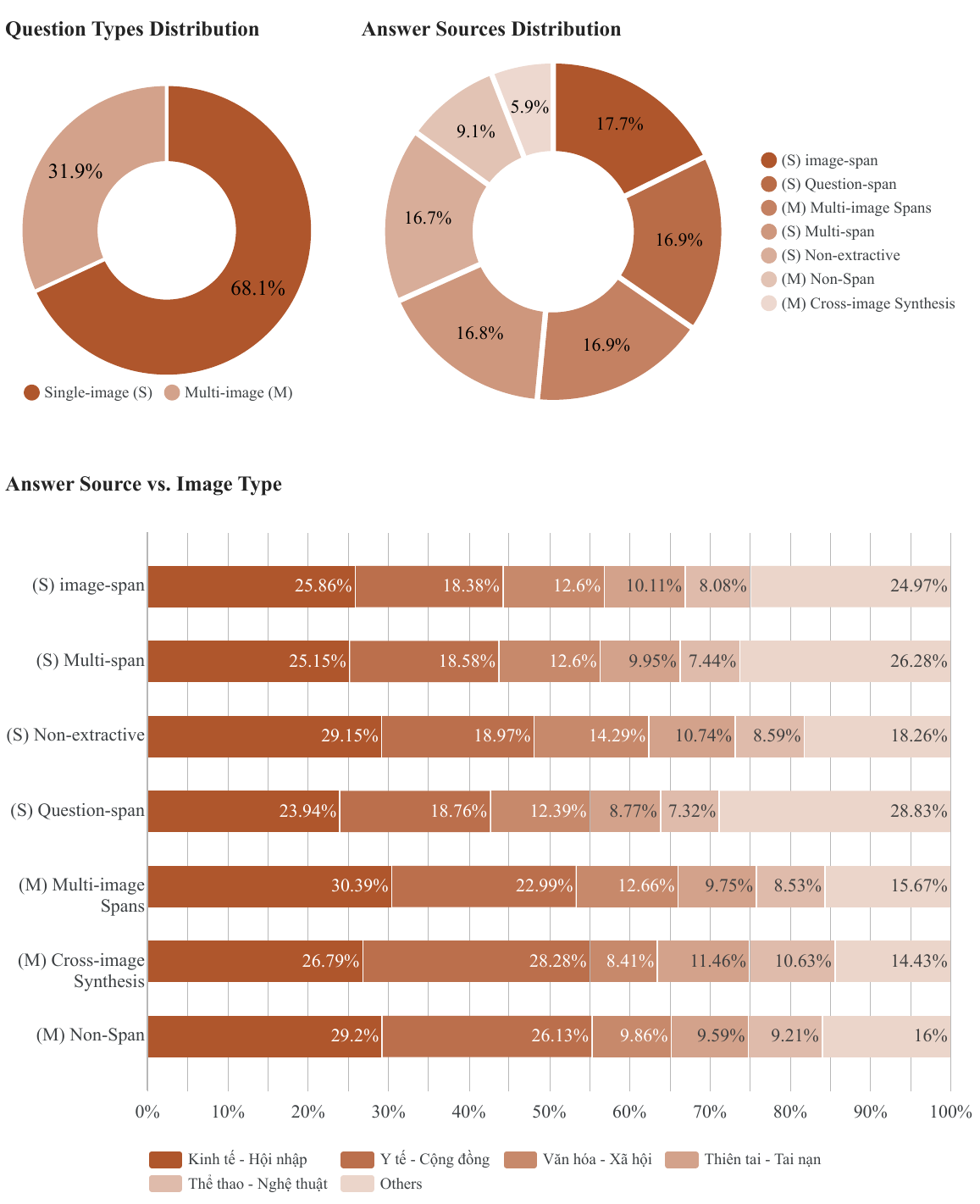}
    \caption{Distribution of topics per source in ViInfographicVQA. The inner ring represents answer sources, while the outer ring shows topic distribution for each source.}
    \label{fig:data_statistics}
\end{figure}

\subsection{Data Statistics}
Figure~\ref{fig:data_statistics} reports the empirical distribution over answer-source categories and topics. \SingleImage questions account for 68.1\% of QAs, while \MultiImage questions comprise 31.9\%. The distributions are stratified by topic and answer source during split construction (details in Table~\ref{tab:accounting}). This balanced composition demonstrates the dataset’s comprehensive coverage of reasoning strategies, spanning from direct extraction to non-extractive inference and cross-image synthesis.

The joint distribution of answer source and image type further reveals that, within the \SingleImage setting, \ImageSpan, \QuestionSpan, \MultiSpan, and \NonExtractive categories appear with comparable frequencies, whereas in the \MultiImage setting, \MultiImageSpan account for the largest proportion, followed by \NonSpan and \CrossImageSynthesis, highlighting the dataset’s emphasis on reasoning that integrates complementary evidence across multiple images.

Across both settings, the top-5 image categories are Economy-Integration, Healthcare-Community, Culture-Society, Disaster-Accident, and Sports-Arts. These domains constitute the majority of the dataset. The remaining categories, grouped as miscellaneous, contribute smaller proportions but enhance the diversity and real-world coverage of the dataset.

\subsection{Dataset Accounting and Splits}
We provide auditable counts for each subset and enforce set-level isolation for \MultiImage groups to avoid cross-split leakage. The data are stratified by topic and answer-source to maintain balanced representation across domains and question types. Additionally, Table~\ref{tab:accounting} summarizes the number of images and question–answer pairs in each subset of \textbf{ViInfographicVQA}. We include shared images infographics that occur in both the \SingleImage and \MultiImage configurations within the same split to enable cross-task evaluation without inducing train-test leakage. This structured organization ensures reproducible evaluation and enables fair comparison between the \SingleImage and \MultiImage tasks.

% We provide auditable counts for each subset and enforce set-level isolation for \MultiImage groups to avoid cross-split leakage. The data are stratified by topic and answer-source to maintain balanced representation across domains and question types. Additionally, Table~\ref{tab:accounting} summarizes the number of images and question–answer pairs in each subset of \textbf{ViInfographicVQA}. This structured organization ensures reproducible evaluation and enables fair comparison between the \SingleImage and \MultiImage tasks.

\begin{table}[t]
\centering
\caption{Dataset statistics and splits of \textbf{ViInfographicVQA}.}
\label{tab:accounting}
\setlength{\tabcolsep}{8pt}
\begin{tabular}{@{}lrrr@{}}
\toprule
\textbf{Subset} & \textbf{\#Images} & \textbf{\#Groups} & \textbf{\#QAs} \\
\midrule
\SingleImage{} (train) & \NStr{} & -- & \QStr{} \\
\SingleImage{} (test)  & \NSte{} & -- & \QSte{} \\
\MultiImage{} (train)  & \NMtr{} & \NMGrpTr{} & \QMtr{} \\
\MultiImage{} (test)   & \NMte{} & \NMGrpTe{} & \QMte{} \\
% \midrule
% Shared image (train) & \NOverlapTr{} & -- & -- \\
% Shared image (test)  & \NOverlapTe{} & -- & -- \\
\midrule
Total train & \textbf{\the\numexpr \NStr+\NMtr-\NOverlapTr\relax} 
             & \textbf{\the\numexpr \NMGrpTr\relax} 
             & \textbf{\the\numexpr \QStr+\QMtr\relax} \\
Total test  & \textbf{\the\numexpr \NSte+\NMte-\NOverlapTe\relax} 
             & \textbf{\the\numexpr \NMGrpTe\relax} 
             & \textbf{\the\numexpr \QSte+\QMte\relax} \\
\bottomrule
\end{tabular}
\end{table}

\begin{table*}[h]
\centering
\caption{Performance of VLMs on the \SingleImage task, categorized by answer type, element, and operation.}
\label{tab:singleimage_performance}
\resizebox{\textwidth}{!}{%
\begin{tabular}{@{}l
c
*{4}{c}
*{6}{c}@{}}
\toprule
Method & Overall
& \multicolumn{4}{c}{Answer Type}
& \multicolumn{6}{c}{Element} \\
\cmidrule(lr){3-6} \cmidrule(lr){7-12}
&
& \makecell[c]{Image\\span}
& \makecell[c]{Multi\\span}
& \makecell[c]{\NonExtractive}
& \makecell[c]{Question\\span}
& \makecell[c]{Diagram/\\Graph}
& Map
& Sequence
& Table
& Text
& \makecell[c]{Visual/\\Layout} \\
\midrule
Phi-4-multimodal-5B & 41.91 & 45.36 & 23.90 & 39.83 & 58.84 & 42.87 & 43.29 & 41.36 & 46.46 & 42.11 & 39.23 \\
VideoLLaMA3 Image-7B & 49.60 & 55.38 & 35.52 & 44.16 & 63.39 & 55.96 & 49.55 & 52.48 & 60.01 & 49.01 & 40.83 \\
InternVL3.5-8B & 67.02 & 73.31 & 49.30 & \best{65.70} & 79.76 & 73.06 & 66.84 & \second{70.82} & 67.56 & 64.30 & \second{64.04} \\
MiniCPM-o2.6-8B & 57.03 & 62.52 & 44.63 & 53.76 & 67.20 & 60.39 & 53.75 & 61.86 & 60.17 & 55.94 & 54.11 \\
Molmo D-7B & 38.26 & 46.66 & 15.89 & 29.19 & 61.48 & 41.44 & 39.19 & 32.48 & 46.08 & 38.94 & 34.38 \\
Ovis2.5-9B & \best{71.02} & \best{78.19} & \best{61.42} & 61.24 & \best{83.21} & \best{74.81} & 67.02 & \best{72.77} & \best{77.45} & \best{69.83} & \best{68.56} \\
Qwen2.5-VL-7B & 67.42 & \second{73.54} & 50.39 & \second{65.63} & 80.14 & \second{73.54} & \best{68.81} & 65.95 & \second{75.38} & 66.25 & 61.59 \\
\midrule
Qwen2.5-VL-7B (Finetuned) & \second{67.80} & 72.69 & \second{53.89} & 63.93 & \second{81.10} & 72.95 & \second{67.13} & 69.35 & 73.65 & \second{68.84} & 61.36 \\
\bottomrule
\end{tabular}%
}
\end{table*}

\section{Experiments}
In this section, we evaluate the performance of several state-of-the-art VLMs on \textbf{ViInfographicVQA}. The experiments measure both zero-shot reasoning capabilities and the performance gains achieved through supervised fine-tuning, providing a holistic analysis of the challenges posed by our dataset.

\subsection{Evaluation Metrics}
We adopt Average Normalized Levenshtein Similarity (ANLS) \cite{biten2019scene}, which offers a flexible evaluation by tolerating minor textual deviations. It is computed as follows:
\begin{equation}
\text{ANLS} = \frac{1}{N} \sum_{i=1}^{N} \max_{j} s(a_{ij}, o_{q_i})
\end{equation}
where $N$ is the number of questions, $o_{q_i}$ is the ground-truth answer for question $i$, and $a_{ij}$ denotes the $j$-th predicted answer for that question. The similarity function $s$ is defined as:
\begin{equation}
s(a_{ij}, o_{q_i}) =
\begin{cases}
1 - \text{NL}(a_{ij}, o_{q_i}) & \text{if } \text{NL}(a_{ij}, o_{q_i}) < 0.5 \\
0 & \text{otherwise}
\end{cases}
\end{equation}
where $\text{NL}(\cdot)$ is the normalized Levenshtein distance. ANLS credits near-matches while penalizing semantically incorrect answers. The final score ranges from 0 to 1, where a higher value indicates better performance by reflecting a closer match between predicted and ground-truth answers. This metric is widely used in text-centric document VQA~\citep{mathew2021docvqa}.

\subsection{Experimental Results}
We benchmark recent open-source VLMs, including a fine-tuned variant of Qwen2.5-VL-7B, to assess the impact of supervised training.\footnote{Models were fine-tuned via QLoRA ($r=8$, $\alpha=32$, dropout $0.05$) for 5 epochs with learning rate $2\times10^{-4}$.}
For the \SingleImage task, we evaluate InternVL3.5-8B \cite{wang2025internvl3}, MiniCPM-o2.6-8B \cite{yao2024minicpm}, Molmo D-7B \cite{deitke2025molmo}, Ovis2.5-9B \cite{lu2025ovis2}, Phi-4-multimodal-5B \cite{abouelenin2025phi}, Qwen2.5-VL-7B \cite{bai2025qwen2}, and VideoLLaMA3 Image-7B \cite{damonlpsg2025videollama3}. For the \MultiImage task, we analyze InternVL3.5-8B, MiniCPM-o2.6-8B, Phi-4-multimodal-5B, and Qwen2.5-VL-7B. Results are summarized in Tables~\ref{tab:singleimage_performance} and \ref{tab:multiimage_performance}.

\begin{table}[h]
\centering
\caption{Model performance (\%) on the \MultiImage dataset by question type.}
\label{tab:multiimage_performance}
\resizebox{\linewidth}{!}{%
\begin{tabular}{@{}lcccc@{}}
\toprule
Method & Overall & \CrossImageSynthesis & \MultiImageSpan & \NonSpan \\
\midrule
InternVL3.5-8B & 35.50 & 10.97 & 51.72 & 24.67 \\
MiniCPM-o2.6-8B & 40.55 & 35.22 & 48.74 & 30.05 \\
Phi-4-multimodal-5B & 20.95 & 20.93 & 24.59 & 14.60 \\
Qwen2.5-VL-7B & \underline{54.92} & \underline{56.34} & \underline{58.33} & \textbf{47.96} \\
\midrule
Qwen2.5-VL-7B (Finetuned) & \textbf{55.47} & \textbf{56.55} & \textbf{58.91} & \underline{47.68} \\
\bottomrule
\end{tabular}%
}
\end{table}

\paragraph{\SingleImage}
Table~\ref{tab:singleimage_performance} shows that Ovis2.5-9B achieves the highest overall ANLS, followed by InternVL3.5-8B and Qwen2.5-VL-7B. Span-based categories (\ImageSpan, \QuestionSpan) remain the easiest, often exceeding 75\% ANLS, while \NonExtractive reasoning is consistently lower across models. Ovis2.5-9B leads on \MultiSpan\ and most element types, particularly tables and graphs, whereas InternVL3.5-8B performs best on \NonExtractive\ reasoning. Fine-tuning yields moderate gains for Qwen2.5-VL, slightly improving arithmetic and reasoning-oriented categories. MiniCPM-o2.6-8B performs mid-range, and lighter models such as Phi-4-multimodal-5B and Molmo D-7B trail far behind. Overall, the results indicate that while modern VLMs handle span extraction reliably, numerical and multi-step reasoning over structured layouts remain significant challenges.

\paragraph{\MultiImage}
Across models, \MultiImage performance drops substantially relative to \SingleImage, typically by about 12–32 ANLS points, underscoring the added difficulty of reasoning over multiple images. Qwen2.5-VL-7B (Finetuned) is the clear leader overall, with the strongest results on \CrossImageSynthesis\ and \MultiImageSpan, while the base Qwen2.5-VL-7B attains the best \NonSpan\ score. Qwen2.5-VL-7B (base) ranks second overall. MiniCPM-o2.6-8B and InternVL3.5-8B follow, with InternVL3.5-8B ranking third on \MultiImageSpan\ but weak on \CrossImageSynthesis. Lighter or earlier-generation models such as Phi-4-multimodal-5B continue to lag, highlighting that robust cross-image synthesis remains challenging for current VLMs.
% Across models, \MultiImage score drops by approximately 14-22 ANLS points versus \SingleImage, with Qwen2.5-VL-7B leading at 52.38\% overall and 45.79\% on \NonSpan. Qwen2.5-VL-7B also attains the highest scores on \CrossImageSynthesis and \MultiImageSpan (both $>54\%$). InternVL3.5-8B ranks second overall (44.75\%) and on span-based aggregation, while MiniCPM-o2.6-8B is the runner-up for \NonSpan numerical reasoning (32.49\%). The gap to lighter or earlier-generation models (e.g., Phi-4-multimodal-5B at 21.75\%) underscores that robust cross-image synthesis remains challenging for current VLMs.

% \paragraph{Evaluation Protocol.}
% Unless noted, decoding uses greedy search with max generation length 64 tokens and an input resolution of the model’s default shortest side (typically 448-1024 px); OCR (if used by a model) is the model’s native pipeline. We report ANLS on the official test split with fixed IDs; no test-time augmentation is applied.

\subsection{Ablation Study}
We further compare supervision strategies: (1) zero-shot inference, (2) fine-tuning on the \SingleImage subset, (3) fine-tuning on the \MultiImage subset, and (4) fine-tuning on the combined single- and \MultiImage data. Table~\ref{tab:fine_tuning_results} indicates that fine-tuning consistently improves ANLS over inference. Fine-tuning on the \MultiImage subset particularly boosts cross-image reasoning, while joint training achieves the best overall performance, suggesting that heterogeneous supervision strengthens both intra-document and cross-image reasoning.

\begin{table}[h]
\centering
\caption{Performance (\%) of Qwen2.5-VL (7B) under different fine-tuning configurations.}
\label{tab:fine_tuning_results}
\resizebox{\linewidth}{!}{%
\begin{tabular}{@{}lccc@{}}
\toprule
Method & \SingleImage & \MultiImage & Combined \\
\midrule
Zero-shot Inference & 67.42 & 54.92 & 63.43 \\
Fine-tuning (\SingleImage only) & \second{67.80} & 54.74 & \second{63.64} \\
Fine-tuning (\MultiImage only) & 67.30 & \second{55.47} & 63.52 \\
Fine-tuning (Combined) & \best{68.53} & \best{55.62} & \best{64.41} \\
\bottomrule
\end{tabular}%
}
\end{table}

These findings underscore the complementary roles of single- and multi-image supervision in multimodal learning, where fine-tuning on single-image data primarily enhances localized visual–textual comprehension, while multi-image supervision facilitates cross-image relational reasoning. Combining both approaches yields the best overall performance, highlighting the importance of balanced, heterogeneous supervision for comprehensive infographic understanding.

\section{Discussion}
Our study provides a realistic and challenging setting for Vietnamese infographic VQA by requiring models to coordinate text understanding, layout structure, and visual evidence, often across multiple related images. Our results indicate that contemporary VLMs handle extractive, \SingleImage questions comparatively well, yet they struggle when arithmetic, multi-step synthesis, or cross-image aggregation is required. This gap suggests that current pipelines, which typically rely on shallow OCR-language fusion, lack explicit mechanisms for layout-aware reasoning and information integration across documents (e.g., memory over sets, retrieval and alignment, or planning-based composition).

A second challenge concerns evaluation. While ANLS is robust to minor orthographic variation, it does not fully capture semantic correctness, numerical exactness, grounding faithfulness, or calibration, limitations that are especially salient for non-extractive and \MultiImage questions. Future evaluations should complement ANLS with task-specific measures (e.g., exact/relative numeric accuracy, unit normalization), faithfulness/attribution checks, and targeted human assessment for ambiguous cases. Finally, although our dataset covers diverse topics within Vietnamese infographics, extending to multilingual and cross-lingual settings, auditing source/template biases, and probing domain transfer remain important next steps for assessing generalization of multimodal systems.

% \section{Conclusion}
% We introduced \textbf{ViInfographicVQA}, the first Vietnamese benchmark for infographic VQA that jointly evaluates \SingleImage comprehension and \MultiImage (cross-image) reasoning, curated from real-world infographics under explicit geometry constraints, organized into semantically coherent sets, and annotated via a VLM-assisted, human-verified pipeline with standardized answer-source categories; we further enforce set-level isolation and stratified splits to support robust, reproducible evaluation. Experiments with recent open-source VLMs show mid-60\% ANLS on \SingleImage questions, especially for image/question-span answers, yet a substantial drop of about 14-22 ANLS points for \MultiImage and non-extractive reasoning, highlighting cross-image synthesis as a key bottleneck; ablations indicate that targeted fine-tuning on \MultiImage supervision improves cross-image performance, and joint training on single- and \MultiImage data yields the best overall trade-off. We hope this benchmark catalyzes progress on layout-aware modeling, tool-augmented arithmetic and table/chart reasoning, data-efficient adaptation without rationale leakage, and richer evaluation protocols that complement ANLS with faithfulness and calibration measures for reliable, real-world multimodal reasoning.

\section{Conclusion}
We construct \textbf{ViInfographicVQA}, the first Vietnamese benchmark for InfographicVQA that jointly evaluates \SingleImage comprehension and \MultiImage (cross-image) reasoning. The dataset is curated from real-world infographics under explicit geometry constraints, organized into semantically coherent sets, and annotated via a VLM-assisted, human-verified pipeline with standardized answer-source categories; we further enforce set-level isolation and stratified splits to support robust, reproducible evaluation. Experiments with recent open-source VLMs show mid-60\% ANLS on \SingleImage questions, especially for image-/question-span answers, yet a substantial drop of about 12-32 ANLS points for \MultiImage and non-span reasoning, highlighting cross-image synthesis as a key bottleneck. Ablation results indicate that targeted fine-tuning on \MultiImage supervision improves cross-image performance, while joint training on both \SingleImage and \MultiImage data yields the best overall trade-off. We hope this benchmark catalyzes progress on layout-aware modeling, tool-augmented arithmetic and table/chart reasoning, data-efficient adaptation without rationale leakage, and richer evaluation protocols that complement ANLS with faithfulness and calibration measures for reliable, real-world multimodal reasoning.

% \section{Acknowledgments}
% This research was fully supported by AI VIETNAM.

\bibliography{aaai2026}
\newpage
\appendix
\onecolumn
\section{Appendix A. Prompt Templates}
To ensure reproducibility, we provide the exact prompt templates used for VLM-assisted QA pair generation. These include rule-based templates for \SingleImage tasks and structured templates for \MultiImage tasks.

\begin{table}[!htbp]
\small
\setlength{\tabcolsep}{6pt}
\begin{tabular}{@{}p{0.28\textwidth} p{0.7\textwidth}@{}}
\toprule
\textbf{Category} & \textbf{Prompt Template} \\
\midrule

\SingleImage -- \textit{Image-span} &
You are given an infographic image. Your task is to generate 2--3 factual question--answer (QA) pairs whose answers are verbatim spans visible in the image.\par
Your answer should be a single word, number, or phrase exactly as it appears in the infographic. Do not infer or paraphrase.
The output format should be \texttt{['Answer']}\par
Remember to generate the final answer only without any additional text.\par
Image: \texttt{<image>} \\

\midrule
\SingleImage -- \textit{Question-span} &
You are given an infographic image. Your task is to generate 2–3 multiple-choice question–answer (QA) pairs where each question explicitly lists several options within the question text.\par
The answer must be exactly one option, verbatim, from the provided list of options. Do not paraphrase or infer beyond what is written.
The output format should be \texttt{['Answer']}.\par
Make sure each question lists explicit options, the answer matches exactly one of them without paraphrasing or multiple selections, and output only the final answer without any extra text.\par
Image: \texttt{<image>} \\

\midrule
\SingleImage -- \textit{Multi-span} &
You are given an infographic image. Your task is to generate 2–3 question–answer (QA) pairs where each answer consists of two or more non-contiguous text segments, all appearing verbatim in the infographic.\par
The question should be a single sentence that clearly instructs how to select these segments (not two separate questions).
Each text segment in the answer must be copied exactly as it appears in the image, without merging, paraphrasing, or adding connective words.\par
The output format should be an array of strings or a single string with a consistent delimiter (e.g., \texttt{['Answer1', 'Answer2']}).\par
Remember to generate the final answer only without any additional text.\par
Image: \texttt{<image>} \\

\midrule
\SingleImage -- \textit{Non-extractive} &
You are given an infographic image. Your task is to generate 2–3 question–answer (QA) pairs where each answer is not directly extractable from the image but must be computed or inferred using the information shown.\par
The final answer must be a single numeric value (integer, decimal, or percentage). Do not provide textual explanations or multiple values. 
Ensure that the question requires reasoning or calculation and that the final value is not explicitly written in the infographic.\par
The output format should be \texttt{['Numeric answer']}.\par
Remember to generate the final numeric answer only without any additional text.\par
Image: \texttt{<image>} \\

\midrule
\MultiImage -- \textit{\MultiImageSpan} &
You are given a set of related infographic images. Write 2 QAs whose answers combine textual spans from at least two images under a logical condition.\par
Your answer should list the merged spans exactly as they appear across images. Use brackets for multiple items: \texttt{['Answer1','Answer2']}. Do not explain.\par
Remember to generate the final answer only without any additional text.\par
Images: \texttt{<image1>, <image2>} \\

\midrule
\MultiImage -- \textit{Cross-image synthesis} &
You are given a set of related infographic images. Write 2 QAs that require comparing, aligning, or interpreting information across images to reach a conclusion.\par
Your answer should be concise and directly supported by the visuals. Include units as shown if applicable. Do not generate reasoning steps or explanations.\par
Remember to generate the final answer only without any additional text.\par
Images: \texttt{<image1>, <image2>} \\

\midrule
\MultiImage -- \textit{Non-span numeric} &
You are given a set of related infographic images. Write 2 QAs requiring numerical operations such as sum, difference, or average computed across the images.\par
Your answer should be a single number or short phrase with the exact unit shown (e.g., \emph{million}, \emph{B}, \emph{K}). Do not show calculation steps.\par
Remember to generate the final answer only without any additional text.\par
Images: \texttt{<image1>, <image2>} \\

% \midrule
% Verification&
% You are given a question, its predicted answer, and the corresponding image(s). Verify whether the answer is fully supported by the image content.\par
% Your output should be either \texttt{valid} or \texttt{invalid}. Do not include explanations or justifications.\par
% Remember to generate the final label only without any additional text.\par
% Input: \texttt{<question, answer, image(s)>} \\

\bottomrule
\end{tabular}
\caption{Prompt templates used for VLM-assisted QA generation and verification across \SingleImage{} and \MultiImage{} tasks.}
\end{table}

\end{document}